%% file: ICML_source.tex
\documentclass{article} 

\usepackage{hyperref}       
\usepackage{url}            
\usepackage{booktabs}    
\usepackage{amsfonts}       
\usepackage{nicefrac}       
\usepackage{microtype}      
\usepackage{xspace}
\usepackage{subfigure}

\usepackage{verbatim}

\usepackage{xcolor}
\usepackage{amsmath,amsthm,amssymb}
\usepackage{mathtools}
\usepackage{tikz}
\newtheorem{theorem}{Theorem}[section]
\usepackage{dsfont}
\DeclareMathAlphabet{\mathbbold}{U}{bbold}{m}{n}

 \usepackage[cal=dutchcal,
 calscaled=1,
 bb=boondox,
 scr=euler]{mathalfa}
\usepackage{wrapfig}
\usepackage{verbatim}
\usepackage[capitalise]{cleveref}
\crefformat{equation}{(#2#1#3)}

\input{math_commands.tex}

\usepackage{hyperref}
\usepackage{url}

\usepackage[accepted]{icml2020}

\surroundwithmdframed[
  leftmargin=1pt,
  skipabove=\medskipamount,
  skipbelow=\medskipamount
]{example}

\begin{document}

\twocolumn[
\icmltitle{Universal Equivariant Multilayer Perceptrons}

\begin{icmlauthorlist}
\icmlauthor{Siamak Ravanbakhsh}{mcgill,mila}
\end{icmlauthorlist}

\icmlaffiliation{mcgill}{School of Computer Science, McGill University, Montreal Canada.}
\icmlaffiliation{mila}{Mila - Quebec AI Institute.}
\icmlcorrespondingauthor{Siamak Ravanbakhsh}{siamak@cs.mcgill.ca}

\icmlkeywords{Equivariant MLP, MLP, Universal Approximation, Invariance, Equivariance, Symmetry, Neural Networks, Deep Learning, Equivariant Networks, Group CNN}

\vskip 0.3in
]

\printAffiliationsAndNotice{}  


\begin{abstract}
Group invariant and equivariant Multilayer Perceptrons (MLP), also known as Equivariant Networks, have achieved remarkable success in learning on a variety of data structures, such as sequences, images, sets, and graphs. 
Using tools from group theory, this paper proves the universality of a broad class of equivariant MLPs with a single hidden layer.  
In particular, it is shown that having a hidden layer on which the group acts \emph{regularly} is sufficient for universal equivariance (invariance). A corollary is unconditional universality of equivariant MLPs for Abelian groups, such as CNNs with a single hidden layer.
A second corollary is the universality of equivariant MLPs with a high-order hidden layer, where we give both group-agnostic bounds and means for calculating group-specific bounds on the order of hidden layer that guarantees universal equivariance (invariance). 
\end{abstract}

\section{Introduction}
Invariance and equivariance properties constrain the output of a function under various transformations of its input. This constraint serves as a strong  learning bias that has proven useful in sample efficient learning for a wide range of structured data. 
In this work, we are interested in universality results for Multilayer Perceptrons (MLPs) that are constrained to be equivariant or invariant. 
This type of result guarantees that the model can approximate any continuous equivariant (invariant) function with an arbitrary precision, in the same way an unconstrained MLP can approximate an arbitrary continuous function~\cite{hornik1989multilayer,cybenko1989approximation,funahashi1989approximate}. 

Study of invariance in neural networks goes back to the book of Perceptrons~\cite{minsky2017perceptrons}, 
where the necessity of parameter-sharing for invariance was used 
to prove the limitation of a single layer Perceptron. 
The follow-up work  showed how parameter symmetries can be used to achieve invariance to finite and infinite groups~\cite{shawe1989building,wood1996representation,shawe1993symmetries,wood1996invariant}. These fundamental early works went unnoticed during the resurgence of neural network research and 
renewed attention to symmetry~\cite{hinton2011transforming,mallat2012group,bruna2013invariant,gens2014deep,jaderberg2015spatial,dieleman2016exploiting,cohen2016group}.

 When equivariance constraints are imposed on feed-forward layers in an MLP, the linear maps in each layer is constrained to use tied parameters~\cite{wood1996representation,ravanbakhsh_symmetry}. This model that we call an \emph{equivariant MLP} appears in  deep learning with sets~\citep{zaheer_deepsets,qi2017pointnet}, exchangeable tensors~\cite{hartford2018deep}, graphs \cite{maron2018invariant}, and relational data~\citep{graham2019deep}. Universality results for some of these models exists~\citep{zaheer_deepsets,segol2019universal,keriven2019universal}. Broader results for 
 high order \emph{invariant} MLPs appears in \citep{maron2019universality}; see also \citep{yarotsky2018universal}. 

A parallel line of work in equivariant deep learning studies linear action of a group beyond permutations.
The resulting equivariant linear layers can be written using convolution operations~\cite{cohen2016steerable,kondor2018generalization}.
When limited to permutation groups, group convolution is simply another 
expression of parameter-sharing~\cite{ravanbakhsh_symmetry}; see also \cref{sec:conv}.
However, in working with linear representations, one may move beyond finite groups~\cite{cohen2019general}; see also~\cite{wood1996representation}.
Some applications include equivariance to isometries of the Euclidean space~\citep{weiler2019general,worrall2017harmonic}, and sphere~\citep{cohen2018spherical}. 
Extension of this view to manifolds is proposed in~\cite{cohen2019gauge}. 
Finally, a third line of work in equivariant deep learning that involves a specialized architecture and learning procedure is that
of Capsule networks~\citep{sabour2017dynamic,hinton2018matrix}; 
see ~\cite{lenssen2018group} for a group theoretic generalization.

\subsection{Summary of Results}
This paper proves universality of equivariant MLPs for finite groups in several settings:
Our main theorems show that any equivariant MLP with a single \emph{regular} hidden layer is universal equivariant (invariant). This has two corollaries: 1)  unconditional universality for Abelian groups, including a two-layer CNN; 2) universality of equivariant MLPs with a \emph{high-order hidden layer} that subsumes existing universality results for high order networks. More specifically, we prove that a high order hidden layer with an order of $\log(|\H|)$, where $\H$ is the stabilizer group is universal equivariant (invariant).
Using the largest possible stabilizer on a set of size $N$, this leads to a lower-bound smaller than $N \log_2(N)$ for \emph{universal equivariance} to arbitrary permutation group. This bound is an improvement over the previous bound $\frac{1}{2}N(N-1)$ that was shown to guarantee universal ``invariance''~\cite{maron2019universality}.
The second part of the paper more closely examines \emph{product spaces} by decomposing them using \emph{Burnside's table of marks}. Using this tool we arrive at the same group-agnostic bounds above, as well as potentially better group-specific bounds for high-order hidden layers. For example, it is shown that equivariant (hyper) graph networks are universal for the hidden layer of order $N$.

\section{Preliminaries}
Let $\G = \{\g\}$ be a finite group. We define the \emph{action} of this group on two finite sets $\I$ and $\O$ of input and output units in a feedforward layer. Using these actions which define permutation groups we then define equivariance and invariance.
In detail, \G-action on the set $\I$ is a structure preserving map (homomorphism) $\fun{a}: \G \to \S_{\I}$, into the symmetric group $\S_\I$, the group of all permutations of $\I$. The \emph{image} of this map is a permutation group $\G_{\I} \leq \S_{\I}$. Instead of writing $[\fun{a}(\g)](n)$ for $\g \in \G$ and $\i \in \I$, we use the short notation $\g \cdot \i = g^{-1} \i$
to denote this action.\footnote{Using $\g^{-1}$ instead of $\g$ is to make this a \emph{right action} despite appearing on the left hand side of $\i$.}
Let $\O$ be another $\G$-set, where the corresponding permutation action $\G_\O  \leq \S_{\O}$ is defined by $\fun{b}: \G \to \S_{\O}$.
\G-action on $\I$ naturally extends to $\x \in {\Real}^{\I}$ by ${\g \cdot \x}_{\i} \defeq \x_{\g \cdot \i}\; \forall \g \in \G_{\I}.$ 
More conveniently, we also write this action as $\mat{A}_\g \x$,
where $\mat{A}_g$ is the permutation matrix form of $\fun{a}(\g,\cdot):\I \to \I$. 

\subsection{Invariant and Equivariant Linear Maps} 
Let the real matrix $\Phi \in \Real^{|\I| \times |\O|}$  
denote a linear map $\Phi: \Real^{|\I|} \to \Real^{|\O|}$.
We say this map is \G-equivariant iff 
\begin{align}\label{eq:equivariance}
  \mat{B}_\g \Phi \x = \Phi\; \mat{A}_\g \x \quad \forall \x \in \Real^{\I}, \g \in \G.
\end{align}
where similar to $\mat{A}_\g$, the permutation matrix $\mat{B}_\g$ is defined based on the action $\fun{b}(\cdot, \g): \O \to \O$.
In this definition, we assume that the group action on the input is \emph{faithful} -- that is $\fun{a}$ is injective, or $\G_{\I} \cong \G$. If the action on the output index set $\O$ is not faithful, then the \emph{kernel} of this action is a non-trivial \emph{normal subgroup} of $\G$, $\ker(\fun{b}) \triangleleft \G$. In this case $\G_{\O} \cong \G / \ker(b)$ is a \emph{quotient group}, and it is more accurate to say that
$\Phi$ is invariant to $\ker(\fun{b})$ and
equivariant to $\G / \ker(\fun{b})$. Using this convention $\G$-equivariance and $\G$-invariance correspond to extreme cases of $\ker(\fun{b}) = \G$ and $\ker(\fun{b}) = \{e\}$. Moreover, composition of such invariant-equivariant functions preserves this property, motivating design of deep networks by stacking equivariant layers. 

\subsection{Orbits and Homogeneous Spaces}
$\G_{\I}$ partitions $\I$ into \emph{orbits}
$\I_1, \ldots, \I_O$, where $\G_{\I}$
is \emph{transitive} on each orbit, meaning that
for each pair $\i_1, \i_2 \in \I_o$, there is
at least one $\g \in \G_{\I}$ such that $\g \cdot \i_1 = \i_2$. If $\G_\I$ has a single orbit, it is {transitive}, and $\I$ is called a \textit{homogeneous space} for $\G$. If moreover the choice of $\g \in \G_\I$ with $\g \cdot \i_1 = \i_2$ is unique, then $\G_{\I}$
is called \emph{regular}.

Given a subgroup $\H \leq \G$ and $\g \in \G$, the \emph{right coset} of $\H$ in $\G$, defined as $\H\g \defeq \{\h\g, \h \in \H\}$ is
a subset of $\G$. For a fixed $\H \leq \G$, the set of these right-cosets, $\H \backslash \G = \{\H\g, \g \in \G\}$, form a partition of $\G$.
$\G$ naturally acts on the right coset space, where ${\g'} \cdot (\H\g) \defeq \H(\g\g')$ sends one coset to another.
The significance of this action is that ``any'' transitive $\G$-action is isomorphic to $\G$-action on some right coset space.
To see why, note that in this action any $\h \in \H$ \emph{stabilizes} the coset $\H \gr{e}$, because $\h \cdot \H \gr{e} = \H \gr{e}$.\footnote{
More generally, when $\G$ acts on the coset $\H \gr{a} \in \H \backslash \G$, all $\g \in \gr{a}^{-1}\H\gr{a}$ stabilize $\H\gr{a}$. Since $\g = \gr{a}^{-1}h\gr{a}$ for some $\h \in \H$, we have ${(\gr{a}^{-1}\h \gr{a})} \cdot \H\gr{a} = \H(\gr{a} \gr{a}^{-1} \h \gr{a}) = \H \gr{a}$.
This means that any transitive $\G$-action on a set $\I$ may be identified with the stabilizer subgroup $\G_{\i} \defeq \{\g \in \G \; s.t.\; \g \cdot \i = \i\}$, for a choice of $\i \in \I$. 
This gives a bijection between $\I$ and the right coset space $\G_\i \backslash \G$.}
Therefore in any action the stabilizer identifies the coset space.

\subsection{Parameter-Sharing and Group Convolution View}\label{sec:conv}
Consider the equivariance condition of \cref{eq:equivariance}. Since the equality holds for
all $\x \in \Real^{\I}$, and using the fact that the inverse of a permutation matrix is its transpose, the equivariance constraint reduces to
\begin{align}\label{eq:eqvariance_w}
    \mat{B}_{\g} \Phis \mat{A}_{\g}^{\top } = \Phis \quad \forall \g \in \G.
\end{align}
The equation above
ties the parameters within the orbits of $\G$-action on rows and columns of $\Phis$:
\begin{align}\label{eq:param_sharing}
    \Phis ({\o,\i}) = \Phis({\g \cdot \o, \g \cdot \i})  \forall \g \in \G, \i,\o \in \I \times \O
\end{align}
where $\Phis({\g \cdot \o, \g \cdot \i})$ is an element of the matrix $\Phis$. This type of group action on Cartesian product space is sometimes called the \emph{diagonal} action. In this case, the action is on the Cartesian product of rows and columns of $\Phis$.

We saw that any homogenous \G-space is isomorphic to a coset space. Using  
 $\I \cong \H \backslash \G$ and $\O \cong \K \backslash \G$, the parameter-sharing constraint of \cref{eq:eqvariance_w} becomes
 \begin{align}\label{eq:tied_param}
\Phis({\K \g, \H \g'}) &= \Phis({\g^{-1} \cdot \K \g,\g^{-1} \cdot \H \g'})\\ 
&= \Phis({\K, \H \g'\g^{-1}}) \forall \g,\g' \in \G, 
\end{align}
Since we can always multiply both indices to have the coset $\K$ as the first
argument, we can replace the matrix $\Phi$ with the vector $\varphi$, such that
$\Phis(\K \g, \H \g')  = \varphi(\H \g' \g^{-1}) \quad \forall \g, \g' \in \G$.
This rewriting also enables us to express the matrix vector multiplication of the linear map $\Phi$ in the form of cross-correlation of input and a kernel $\varphi$
\begin{align}\label{eq:eq.in.k.kernel}
[\Phis \x](\i) &= [\Phis \x](\K \g) \\
&= \sum_{\H\g' \in \H \backslash \G} \Phis ({\K \g, \H\g'}) \x(\H \g')\\
  & = \sum_{\H\g' \in \H \backslash \G} \varphi (\H\g'\g^{-1}) \x(\H \g') \label{eq:crosscorrelation}
\end{align}
This relates the parameter-sharing view of equivariant maps \cref{eq:tied_param} to the convolution view 
\cref{eq:crosscorrelation}. 
Therefore, the universality results in the following extends to group convolution layers~\citep{cohen2016group,cohen2019general},
for finite groups.

\paragraph{Equivariant Affine Maps} 
We may extend our definition, and consider \emph{affine} \G-maps $\Phis \x + \vect{b}$, by allowing an ``invariant'' \emph{bias} parameter $\vect{b} \in \Real^{|\O|}$ satisfying
\begin{align}\label{eq:equivariance_bias}
\mat{B}_g \vect{b} = \vect{b}.
\end{align}
This implies a parameter sharing constraint
    $\vect{b}(\o) = \vect{b}(\g \cdot \o)$.
For homogeneous $\O$, this constraint enforces 
a \emph{scalar} bias.
Beyond homogeneous spaces, the number of free parameters in $\vect{b}$ grows with the number of orbits.

\subsection{Invariant and Equivariant MLPs}
One may stack multiple layers of equivariant affine maps with multiple channels, followed by a non-linearity, 
so as to build an \emph{equivariant MLP}. One layer of this equivariant MLP \aka \emph{equivariant network} is given by:
\begin{align*}
\x^{(\ell)}_{c} = \sigma \left ( 
    \sum_{c'=1}^{C^{(\ell-1)}} \Phi^{(\ell)}_{{c}, c'} \x^{(\ell-1)}_{c'} + \vect{b}_{c}^{(\ell)} \right ),
\end{align*}
where $1\leq c' \leq C^{(\ell-1)}$ and $1\leq c \leq C^{(\ell)}$ index the input and output channels respectively,
$\x^{(\ell)}$ is the output of layer $1 \leq \ell \leq L$, with $\x^{(0)} = \x$ denoting the original input. Here, we assume that $\G$ faithfully acts on all  $\x^{(\ell)}_c \in \Real^{\set{H}^{(\ell)}} \quad \forall c, \ell$, with $\set{H}^{(0)} = \I$ and $\set{H}^{(L)} = \O$. The parameter matrices $\Phi^{\ell}_{c^{(\ell)}, c^{(\ell)}} \in \Real^{\set{H}^{(\ell-1)} \times \set{H}^{(\ell)}}$, and the bias vector $\vect{b}_c^{(\ell)} \in \Real^{\set{H}^{(\ell)}}$ are constrained by the parameter-sharing conditions \cref{eq:eqvariance_w} and 
\cref{eq:equivariance_bias} respectively.
In an \emph{invariant MLP} the faithfulness condition for  \G-action on the hidden and output layers are lifted. In practice, it is common to construct invariant networks by first constructing an equivariant network followed by pooling over $\set{H}^{(L)}$. 

\begin{figure}
    \centering
    \includegraphics[width=.4\linewidth]{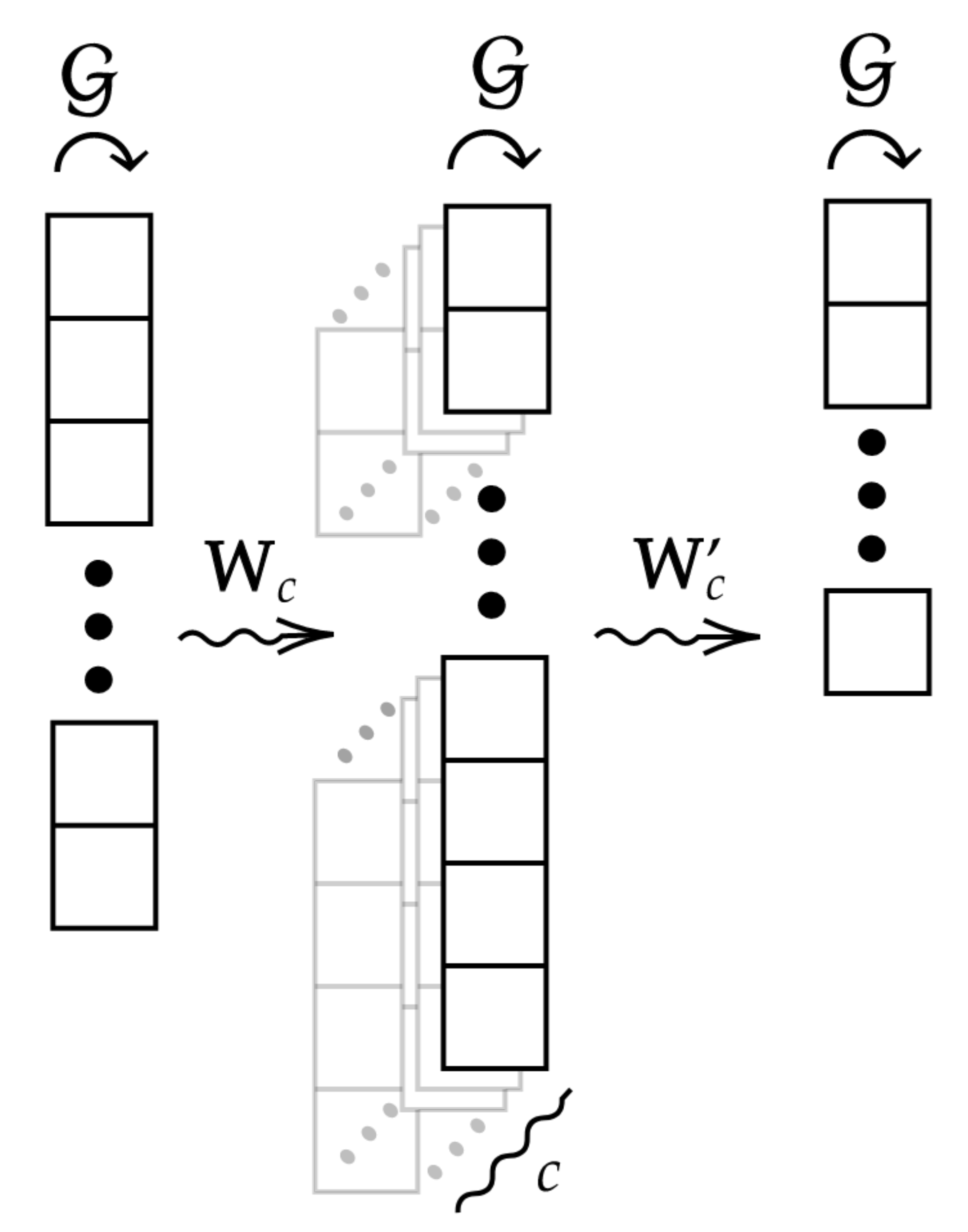}
    \caption{\footnotesize{The equivariant MLP of \cref{eq:universal_equivariant_naive}. The symbol $\curvearrowright$ indicates $\G$-action on the units, $\Phis_c$ and $\Phis'_c$ for all channels of the hidden layer $c = 1,\ldots,C$
    are constrained by the parameter-sharing of \cref{eq:param_sharing}. If $\G$-action on the hidden layer is regular, the number of channels can grow to approximate any continuous $\G$-equivariant function with an arbitrary accuracy. Bias terms are not shown.}}
    \label{fig:equivariant-MLP}
\end{figure}

\section{Universality Results}\label{sec:universality_reg}
This section presents two new results on universality of both invariant and equivariant networks with a single hidden layer ($L=2$). 
Formally, we can claim that a \G-equivariant MLP $\hat{\psi}: \Real^{|\I|} \to \Real^{|\O|}$ is a \emph{universal \G-equivariant approximator}, if for any \G-equivariant continuous function 
$\psi: \Real^{|\I|} \to \Real^{|\O|}$, any compact set $\set{K} \subset \Real^{|\I|}$, and $\epsilon > 0$, 
 there exists a choice of parameters, and number of channels such that $||\psi(\x) - \hat{\psi}(\x)|| < \epsilon \; \forall \x \in \set{K}$.

\begin{mdframed}[style=MyFrame2]
\begin{theorem}\label{th:universal.invariant}
A \G-invariant network  
\begin{align}\label{eq:universal_invariant_naive}
    \hat{\psi}(\x) =  \sum_{c=1}^{C} {w'}_{c}  \mathbf{1}^{\top } \sigma \big (\Phi_{c} \x + b_{c} \big).
\end{align}
with a single hidden layer, on which \G acts regularly is a universal \G-invariant approximator.  
Here, $\mathbf{1} = \underbrace{[1,\ldots,1]^\top }_{|\G|}$ and 
, $b_c,w'_{c} \in \Real$.
\end{theorem}
\end{mdframed}
\begin{proof}
The first step follows the symmetrisization argument~\cite{yarotsky2018universal}.
Since MLP is a universal approximator, for any compact set $\mathds{K} \subset \Real^{|\I|}$, we can find $\psi_{MLP}$ such that for any $\epsilon > 0$, 
$
|\psi(\x) - \psi_{MLP}(\x)| \leq \epsilon 
$ for  $\x \in \mathds{K}$. Let $\mathds{K}_{sym} = \{ \bigcup_{\g \in \G} \mat{A}_\g  \x | \x \in \mathds{K}\}$ denote the symmetrisized $\mathds{K}$, which is again a compact subset of $\Real^\I$ for finite $\G$. Let $\psi_{MLP+}$ approximate $\psi$ on the symmetrisized compact set $\mathds{K}_{sym}$.
It is then easy to show that for \G-invariant $\psi$,
the \emph{symmetrisized MLP} 
$
\psi_{sym}(\x) = \frac{1}{|\G|} \sum_{\g \in \G} \psi_{MLP+}(\mat{A}_\g \x)
$
also approximates $\psi$
\begin{align}
    |\psi(\x) - \psi_{sym}(\x)| = |\psi(\x) - \frac{1}{|\G|}\sum_{\g \in \G} \psi_{MLP+}(\x)| \\
    \leq \frac{1}{|\G|} \sum_{\g \in \G} |\psi(\mat{A}_{\g} \x) - \psi_{MLP}(\mat{A}_{\g} \x)| \leq \epsilon.
\end{align}
Next step, is to show that $\psi_{sym}$ is equal to $\hat{\psi}$ of \cref{eq:universal_invariant_naive}, for some parameters 
 $\Phi_{c} \in \Real^{|\setb{H}| \times |\I|}$ constrained so that $\mat{H}_{\g} \Phi_c  = \Phi_c \mat{A}_\g \forall \g \in \G$, where $\mat{A}_\g$ and $\mat{H}_{\g}$ are the permutation representation of \G action on the input and the hidden layer respectively.
\begin{align}
    \psi_{sym}(\x) &= \frac{1}{|\G|} \sum_{\g \in \G} \sum_{c=1}^{C} w'_c \sigma \big ( \vect{w}_c^{\top } (\mat{A}_\g \x) \big ) \label{eq:p01}\\
    & =  \sum_{c=1}^{C} \frac{w'_c}{|\G|} \sum_{\g \in \G} \sigma \big ( (\vect{w}_c^{\top } \mat{A}_\g) \x \big ) \\
    & = \sum_{c=1}^{C} \tilde{w}_c \mathbf{1}^{\top }\sigma \left ( 
    \underbrace{
    \begin{bmatrix} 
    -\vect{w}_c^{\top } \mat{A}_{\g_1}-\\
    \vdots\\
    -\vect{w}_c^{\top } \mat{A}_{\g_{|\set{H}|}}-
    \end{bmatrix}}_{\Phi_c} \x 
    \right ).
\end{align}
where in the last step we put the summation terms into rows of the matrix $\Phi_c$, and performed the summation using multiplication by $\vect{1}^\top $.
$\tilde{w}_c$ is the rescaled ${w}'_c$.
Since the summation in \cref{eq:p01}  is over $\g \in \G$, each row of $\Phis_c$ and therefore each hidden unit is ``attached'' to exactly one group member, which translates to having a \emph{principal homogeneous space}, \aka a regular \G-set. Note that we have the freedom to choose the rows to have any order, corresponding to a different order in summation,  which means that the choice of a particular principal homogeneous space is irrelevant. 

Now we show that the  parameter matrix $\Phi_c \in \Real^{|\set{H}| \times |\I|}$ 
above satisfy the parameter-sharing constraint 
$\Phi_c \mat{A}_\g = \mat{H}_\g \Phi_c \; \forall \g \in \G$:
\begin{align*}
    \mat{H}_\g \Phi_c \mat{A}_\g^{-1} =  \begin{bmatrix} 
    \vect{w}_c^{\top } \mat{A}_{\g_1 \g}\\
    \vdots\\
    \vect{w}_c^{\top } \mat{A}_{\g_{|\set{H}|} \g}
    \end{bmatrix}
    \mat{A}_{\g^{-1}} 
    = \begin{bmatrix} 
    \vect{w}_c^{\top } \mat{A}_{\g_1}\\
    \vdots\\
    \vect{w}_c^{\top } \mat{A}_{\g_{|\set{H}| }}
    \end{bmatrix} = \Phi_c
\end{align*}
where the first equality follows from the fact that row indexed by $\g_r$ is moved to the row $\g \cdot \g_r = \g_r \g^{-1}$:
$
\mat{H}_{\g} \mat{A}_{\g_r}  = \mat{A}_{\g \cdot \g_r} = \mat{A}_{\g_r \g^{-1}} 
$. Therefore, the current row $\g_{r'}$ was previously $\g^{-1} \cdot \g_{r'} = \g_{r'} \g$.
The second equality follows from $\mat{A}_\g^{-1}$ is acting from the right, and no further inversion is needed
$
\mat{A}_{\g_r \g} \mat{A}^{-1}_{\g}  = \mat{A}_{\g_r \g \g^{-1}} = \mat{A}_{\g_r}.
$
This shows that a \G-invariant network with a single hidden layer on which \G acts regularly is equivalent to a symmetricized MLP, and therefore for some number of channels, it is a universal approximator of \G-invariant functions. 
\end{proof}
This result should not be surprising since the size of a regular hidden layer grows with the group, and as it is evident from the proof, \emph{an equivariant MLP with a regular hidden layer implicitly averages the output over all transformations of the input.}
Next, we apply a similar idea to prove the universality of the \emph{equivariant} MLPs with a regular hidden layer.

\begin{mdframed}[style=MyFrame2]
\begin{theorem}\label{th:universal.equivariant}
A \G-equivariant MLP 
\begin{align}\label{eq:universal_equivariant_naive}
    \hat{\psi}(\x) =  \sum_{c=1}^{C} {\Phi'}_{c}\, \sigma \big ( \Phi_{c} \x + b_{c} \big ).
\end{align}
with a single \emph{regular} hidden layer is a universal \G-equivariant approximator.
\end{theorem}
\end{mdframed}
\begin{proof}
In this setting, symmetricization, using the so-called \emph{Reynolds operator}~\cite{sturmfels2008algorithms}, for the universal MLP is given by
\begin{equation}\label{eq:sym-MLP-equivariant}
    \psi_{sym}(\x) = \frac{1}{|\G|} \sum_{\g \in \G} \mat{B}_{\g^{-1}} \sum_{c=1}^{C} \vect{w}'_c \sigma \big ( \vect{w}_c^{\top } \mat{A}_\g \x + b_c \big)
\end{equation}
where $\vect{w}_c \in \Real^{|\I|}$ and $\vect{w}'_c \in \Real^{\O}$
are the weight vectors in the first and second layer associated with hidden unit $c$.
Our objective is to show that this symmetrisized
MLP is equivalent to the equivariant network of  \cref{eq:universal_equivariant_naive}, in which $\Phi'_{c} \in \Real^{|\O| \times |\set{H}|}$, and $\Phi_{c} \in \Real^{|\set{H}| \times |\I| }$ use parameter-sharing to satisfy 
\begin{align}
    \mat{H}_{\g} \Phi_c  = \Phi_c \mat{A}_\g \; \text{and}\; 
    \mat{B}_{\g} \Phi'_c  = \Phi'_c \mat{H}_\g  \;\forall \g \in \G.
\end{align} 
Here, $\mat{A}_\g$, $\mat{B}_\g$ and $\mat{H}_{\g}$ are the permutation representations of \G action on the input, the output, and the hidden layer respectively.

First, rewrite the symmetrisized MLP as
\begin{align*}
\psi_{sym}(\x) &= \sum_{c=1}^{C} \sum_{\g \in \G} \mat{B}_{\g^{-1}}  \vect{w}'_c \sigma \big ( \vect{w}_c^{\top } \mat{A}_\g \x + b_c \big)\\
    &= \sum_{c=1}^{C} \Phi'_c \sigma \big ( \Phi_c \x \big )\\
    \text{where} \quad \Phi'_c & = \begin{bmatrix}
    | & & | \\
    \mat{B}_{\g_1^{-1}} \vect{w}'_c & \ldots &  \mat{B}_{\g_{|\G|}^{-1}} \vect{w}'_c\\
    | & & | \\
    \end{bmatrix} \\
    \Phi_c & = \begin{bmatrix}
    - \vect{w}_c \mat{A}_{\g_1}- \\ 
    \vdots \\  
    -  \vect{w}_c \mat{A}_{\g_{|\G|}}-
    \end{bmatrix},
\end{align*}
and the $\frac{1}{|\G|}$ factor is absorbed in one of the weights. 
It remains to show that the two matrices above satisfy the equivariance condition $\mat{H}_{\g} \Phi_c  = \Phi_c \mat{A}_\g$ and $\mat{B}_{\g} \Phi'_c  = \Phi'_c \mat{H}_\g$. The proof for $\Phi_c$ is identical to the invariant network case. 

For $\Phi'_c$, we use a similar approach.
\begin{align*}
    \mat{B}_g \Phi'_c \mat{H}_g^{-1} =
    \begin{bmatrix}
    | & & | \\
    \mat{B}_{\g} \mat{B}_{\g_1^{-1} \g} \vect{w}'_c & \ldots &  
    \mat{B}_{\g} \mat{B}_{\g_{|\G|}^{-1}\g} \vect{w}'_c\\
    | & & | \\
    \end{bmatrix} \\
    =
    \begin{bmatrix}
    | & & | \\
     \mat{B}_{\g_1^{-1}} \vect{w}'_c & \ldots &  
     \mat{B}_{\g_{|\G|}^{-1}} \vect{w}'_c\\
    | & & | \\
    \end{bmatrix}
    = \Phi'_c.
\end{align*}
In the first step, since $\mat{H}_{\g}^{-1} = \mat{H}_{\g^{-1}}$ is acting on the right, it moves the column indexed by $\g_l^{-1}$ to $\g_l^{-1} \g^{-1}$.
This means that the column currently at $\g_{l'}^{-1}$ is $\g_{l'}^{-1} \g$.
The second step uses the following:
$
\mat{B}_{\g} \mat{B}_{\g_l^{-1} \g} = \mat{B}_{\g \cdot (\g_l^{-1} \g)} =  
\mat{B}_{\g_l^{-1} \g \g^{-1}} = \mat{B}_{\g_l^{-1}}
$. This, proves the equality of the symmetrisize MLP \cref{eq:sym-MLP-equivariant} to the equivariant MLP of \cref{eq:universal_equivariant_naive}. However, a similar argument to the 
proof of invariant case, shows the universality of $\psi_{sym}$. Putting these together, completes the proof of \cref{th:universal.equivariant}.
\end{proof}

\subsection{Universality for Abelian Groups} \label{sec:abelian}
In the case where \G is an \emph{Abelian group}, any faithful transitive action is regular, meaning that the hidden layer in a \G-equivariant neural network is 
necessarily regular. Combined with \cref{th:universal.equivariant}, this leads to an unconditional universality result for Abelian groups. 
\begin{mdframed}[style=MyFrame2]
\begin{corollary}
 For Abelian group $\G$, 
 a \G-equivariant (invariant) neural network with a single hidden layer is a universal approximator of continuous \G-equivariant (invariant) functions on compact subsets of $\Real^{|\I|}$. 
\end{corollary}
\end{mdframed}
A corollary to this is the universality of a Convolutional Neural Network (CNN) with a single hidden layer. 
\begin{mdframed}[style=MyFrame2]
\begin{corollary}[Universality of CNNs]\label{cor:cnn}
 For an arbitrary input-output dimensions, a CNN with a single hidden layer, full kernels, and cyclic padding is a universal approximator of continuous circular translation equivariant (invariant) functions.
\end{corollary}
\end{mdframed}
Use of the term circular, both in padding and translation is because of the need to work with finite translations, which are produce as the result of the action of a product of cyclic groups.\footnote{Input can be zero-padded, before circular padding, so that \cref{cor:cnn} guarantees universal approximation of translation equivariant functions, where translations are bounded by the size the original input.} 

\subsection{Universality for High-Order Hidden Layers}\label{sec:high-order}
$\G$-action on the hidden units $\set{H}$ naturally extends to its simultaneous action on the Cartesian product $\set{H}^D = \set{H} \times \ldots \times \set{H}$:
$$\g \cdot (h_1,\ldots,h_D) \defeq (\g \cdot h_1,\ldots, \g \cdot h_D).$$
We call this an \emph{order $D$ product space}.
Product spaces are used in building high-order layers in \G-equivariant networks in several recent works~\cite{kondor2018covariant,maron2018invariant,keriven2019universal,albooyeh2019incidence}. 
\citet{maron2019universality} show that for
\begin{align}\label{eq:maron}
D \geq \frac{1}{2} |\set{H}|\;(|\set{H}|-1),
\end{align}
such MLPs with multiple hidden layers of order $D$ become universal \G-\emph{invariant} approximators. 
In this section, we show that better bounds for $D$ that guarantees universal invariance and equivariance follows from the universality results of \cref{th:universal.invariant,th:universal.equivariant}. The next section provides an in-depth analysis of 
product spaces that not only gives an alternative proof of the theorems below, but also could lead to yet better bounds.\footnote{The   beautiful proof for the following theorem was proposed by an anonymous reviewer. The original proof uses the ideas discussed in the next section and appears later in the paper.}
\begin{mdframed}[style=MyFrame2]
\begin{theorem}\label{cor:degree}
 Let $\G$ act faithfully on $\set{H} \cong [\H \backslash \G]$. Then $\set{H}^{D}$ has a regular orbit for any
 $$D \geq \log_2(|\H|)$$
 and therefore, by \cref{th:universal.equivariant}, an order $D$ hidden layer guarantees universal equivariance.
\end{theorem}
\end{mdframed}
\begin{proof}
If $\G$ acts faithfully on $\set{H}$, the intersection of the stabilisers of all the points in $\set{H}$ is trivial -- \ie $\operatorname{Core}_\G(\H) = \{e\}$. If instead of taking the intersection of the stabilisers of all $h \in \set{H}$, we can just take the intersection of the stabilisers of $D$ (carefully chosen) points, we will know there is a regular orbit in $\set{H}^D$. That is because the stabiliser of a point in $\set{H}^d$ is the intersection of the stabilisers of its elements in $\set{H}$, that is $\operatorname{Stab}_\G(h_1,...,h_D)=\bigcap_{d=1}^D \operatorname{Stab}_\G(h_d)$.
So the question is for what value of $D$ can we find $D$ points such that the intersection of their stabilisers is trivial. We work recursively to find a bound on $D$.

Start with just one point $h_1 in \set{H}^1$, and assume its stabiliser is of size $s_1$. Now assume we have a point $(h_1,...,h_d)$ in $\set{H}^d$ such that its stabiliser is of size $s_d$. If $s_d=1$, we are done. Otherwise, since the action is faithful, there has to exist a point $h_{d+1}$ such that the intersection of all the stabilisers of $h_1,...,h_{d+1}$ is a strictly smaller subgroup of the stabiliser of $(h_1,...,h_d)$. The size of a proper subgroup is at most half the size of the original group and therefore $s_{d+1} < s_d / 2$. Therefore, for each additional point the size of stabilizer at least half of the previous stabilizer. It follows that for 
any $D \geq \log_2(|\H|)$, $[\H \backslash \G]^D = \set{H}^D$ has an orbit with a trivial stabilizer.
\end{proof}

Since the largest stabilizer for any action on $\set{H}$ is $\gr{S}_{|\set{H}| -1}$,
we can use a lower-bound for $D$, in \cref{cor:degree} that is independent of the 
stabilizer sub-group $\H$. The following bound follows from the Sterling's approximation 
$ N! < N^{N + \frac{1}{2}} e^{-N+1}$ to the size of the largest possible stabilizer $|\gr{S}_{|\set{H}|-1}| = (|\set{H} - 1|)!$.
\begin{mdframed}[style=MyFrame2]
\begin{corollary}\label{cor:loose}
 The high-order \G-set of hidden units $\set{H}^D$, with $N = |\set{H}|$ has a regular orbit for 
 $$D \geq \lceil (N-\frac{1}{2}) \log_2 (N-1) -(N-2)\log_2(e) \rceil$$
 and following \cref{th:universal.equivariant} the corresponding equivariant MLP is universal approximator
 of continuous $\G$-equivariant functions.
\end{corollary}
\end{mdframed}

\section{Decomposition of Product \G-Sets}\label{sec:product}
A prerequisite to analysis of product \G-sets is their classification,
which also leads to classification of all \G-maps based on their input/output \G-sets.

\subsection{Classification of \G-Sets and \G-Maps}\label{sec:classification}
Recall that any transitive \G-set $\I$ is isomorphic to a right-coset space $\H \backslash \G$. 
However, the right cosets $\H \backslash \G$ and
$(\g^{-1}\H\g) \backslash \G\quad \forall \g \in \G$ are themselves isomorphic. \footnote{
The stabilizer subgroups of two points in a homogeneous space are conjugate, and therefore \G-sets 
resulting from conjugate choice of right-cosets are isomorphic. To see why stabilizers are conjugate, assume $\i = \gr{a}^{-1} \cdot \i$,
and $\h \in \G_{\i}$, then ${\gr{a} \h \gr{a}^{-1}} \cdot \i = \i^{\h \gr{a}} = \i^\gr{a} = \i$. Therefore, ${\gr{a}^{-1} \h \gr{a}} \in \G_{\i}$. Since conjugation is a bijection, this means $\G_{\i} = {\gr{a}^{-1} \G_{\i} a}$.}
This also means what we care about is \emph{conjgacy classes of subgroups} 
$[\H] = \{ \g^{-1} \H \g \mid \g \in \G\},$ 
which classifies right-coset spaces up to conjugacy 
$[\H \backslash \G] = \{ \g^{-1}\H\g \backslash \G \mid \g \in \G\}.$ 
We used the bracket to identify the conjugacy class.
In this notation, for $\H,\H' \leq \G$, we say $[\H] < [\H']$, iff $\g^{-1} \H \g < \H'$, for some $\g \in \G$.

A \G-set is transitive on each of its orbits, and we can identify each orbit with its stabilizer subgroup.
Therefore a list of these subgroups along with their multiplicities completely defines a \G-set up to an isomorphism~\citep{rotman2012introduction}:
\begin{align}\label{eq:disjoint}
\I \cong \bigcup_{[\H_i] \leq \G} p_i [\H_i \backslash \G],
\end{align}
where $p_1,\ldots,p_I \in \set{Z}^{\geq 0}$ denotes the multiplicity of a right-coset space, and $\I$ has $\sum_{i=1}^{I} p_i$ orbits.

To ensure a faithful \G-action on $\I$, 
a necessary and sufficient condition is for the 
point-stabilizers $\G_{\i} \forall \i \in \I$ to have a trivial intersection. The point-stabilizers within each orbit are conjugate to each other and their intersection which is the largest normal subgroup of $\G$ contained in $\H_i$, is called 
the \emph{core} of \G-action on $[\H_i \backslash \G]$:
\begin{align}\label{eq:core}
\mathrm{Core}_\G(\H_i) \defeq \bigcap_{\g \in \G} \g^{-1}  \H_i \g.    
\end{align}

Next, we extend the classification of \G-sets to  \G-equivariant maps, \aka \G-maps $\Phi: \Real^{\I} \to \Real^\O$, by jointly classifying the input and the output index sets $\I$ and $\O$.
We may consider a similar expression to \cref{eq:disjoint} for the output index set 
$
\O = \bigcup_{[\K_j] \leq \G} q_j [\K_j \backslash \G]
$.
The linear \G-map $\Phi: \Real^{\I} \to \Real^{\O}$ is then equivariant to $\G/\K$ and invariant to $\K \triangleleft \G$
iff 
\begin{align}\label{eq:stab_cond}
  \bigcap_{p_i > 0} \operatorname{Core}_\G(\H_i) = \{e\} \; \text{and} \;
 \bigcap_{q_i > 0}  \operatorname{Core}_\G(\K_i) = \K
\end{align}
where the second condition translates to $\K$ invariance of \G-action on $\O$. Note that the first condition is simply ensuring the faithfulness of \G-action on $\I$.
This result means that the multiplicities $(p_1,\ldots,p_I)$ and $(q_1,\ldots,q_J)$ completely identify a (linear) \G-map $\Phi: \Real^{\I} \to \Real^{\O}$ that  equivariant to $\G/\K$ and invariant to $\K \triangleleft \G$, up to an isomorphism.

\subsection{Diagonal Action on Cartesian Product of \G-sets}
Previously we classified all \G-sets as the disjoint union of homogeneous spaces $\bigcup_{i=1}^{I} p_i [\G_i \backslash \G]$, 
where \G acts transitively on each orbit. 
However, as we saw earlier \G also naturally acts on the \emph{Cartesian product} of homogeneous \G-sets:
$$\I_1 \times \ldots \times \I_D = (\G_1 \backslash \G) \times \ldots \times (\G_D \backslash \G)$$
where the action is  defined by
$$\g \cdot (\G_1 \h_1,\ldots, \G_D \h_D) \defeq (\G_1 (\h_1 \g),\ldots, \G_D (\h_D \g)).$$
A special case is when we consider the repeated self-product of the same homogeneous space $\set{H} \cong [\H \backslash \G]$, which as we saw gives an \emph{order $D$ product space}.
$$\set{H}^D \cong [\H \backslash \G]^D = \underbrace{[\H \backslash \G] \times \ldots \times [\H \backslash \G]}_{D \; \text{times}}$$
We call this an \emph{order $D$ product space}.
The following discussion shows how the product space decomposes into orbits, where the existence of a regular orbit leads to universality.


\subsection{Burnside Ring and Decomposition of \G-sets}
Since \emph{any} \G-set can be written as a disjoint union of homogeneous spaces \cref{eq:disjoint}, we expect a decomposition of the product \G-space in the form 
\begin{align}\label{eq:product_space}
[\G_i \backslash \G] \times [\G_j \backslash \G] = \bigcup_{[\G_\ell] \leq \G} \delta^{\ell}_{ i, j} [\G_\ell \backslash \G]
\end{align}
Indeed, this decomposition exists, and the multiplicities $\delta^\ell_{i,j} \in \mathds{Z}^{> 0}$, are called the \emph{structure coefficient} of the \emph{Burnside Ring}.
The (commutative semi)ring structure is due to the fact that the set of non-isomorphic \G-sets
$\Omega(\G) = \{\bigcup_{[\G_i] \leq \G} p_i [\G_i \backslash \G] \mid p_i \in \mathds{Z}^{\geq 0}\},$
is equipped with: 
1) a commutative product operation that is the Cartesian product of \G-spaces, and; 2) a summation operation that is the disjoint union of \G-spaces~\cite{dieck2006transformation}.
A key to analysis of product \G-spaces is finding the structure coefficients in \cref{eq:product_space}.

\begin{example}[\textsc{Product of Sets}]
The symmetric group $\gr{S}_\I$ acts faithfully on $\I$, where the stabilizer is
$\gr{S}_\i = \gr{S}_{\I - \{\i\}}$ -- that is the stabilizer of $\i \in \I $ is the set of all permutations of the remaining items $\I - \{\i\}$. This means
$\I \cong [\gr{S}_{\I - \{\i\}} \backslash \gr{S}_\I]$.

The diagonal $\gr{S}_\I$ action on the product space $\I^D$,
decomposes into $\sum_{i} p_i = \mathrm{Bell}(D)$ orbits, where the Bell number is the number of different partitions of a set of $D$ labelled objects~\cite{maron2018invariant}. 
One may further refine these orbits by their type in the form of \cref{eq:product_space}: 
\begin{align}\label{eq:product.set.decomp}
    [\gr{S}_{\I - \i} \backslash \gr{S}_\I]^D = \bigcup_{d=1}^{D} \mathrm{S}(D, d)[\gr{S}_{\I - \{\i_1,\ldots,\i_d\}} \backslash \gr{S}_\I]
\end{align}
where the ``structure coefficient'' $\mathrm{S}(D, d)$ is the \emph{Stirling number of the second kind}, and it counts the number of ways $D$ could be partitioned into $d$ non-empty sets.
For example, when $D = 2$, one may think of the index set $\I \times \I$ 
as indexing some $|\I| \times |\I|$ matrix. This matrix decomposes into one  ($\mathrm{S}(2,1)=1$) diagonal $[\gr{S}_{\I - \{\i\}} \backslash \gr{S}_\I]$ and one
$\mathrm{S}(2,2) = 1$ set of  off-diagonals $[\gr{S}_{\I - \{\i_1, \i_2\}} \backslash \gr{S}_\I]$. 
This decomposition is presented in \cite{albooyeh2019incidence}, where it is shown that these orbits correspond to ``hyper-diagonals''
for higher order tensors.
For general groups, inferring the structural coefficients is
more challenging, as we see shortly.
\end{example}

From \cref{eq:product.set.decomp} in the example above it follows that an order $D=|\I|$ product of sets contains a regular orbit. The following is a corollary that combines this with the universality results of \cref{th:universal.invariant,th:universal.equivariant}. 
\begin{mdframed}[style=MyFrame2]
\begin{corollary}\label{cor:set.universality}[Universality of Equivariant Hyper-Graph Networks]
A $\gr{S}_\I$ equivariant network with a hidden layer of order $D \geq {|\I|}$, is a universal approximator of $\gr{S}_\I$-equivariant (invariant) functions, where the input and output layer may be of any order. 
\end{corollary}
\end{mdframed}
Note how using group specific analysis gives a better bound of $D \geq N$ compared to group agnostic bound $D \geq N \log(N)$ of \cref{cor:loose}.
A universality result for the invariant case only, using a quadratic order appears in \cite{maron2019universality}, where the MLP is called a \emph{hyper-graph network}. 
\citet{keriven2019universal} prove universality for the equivariant case, without giving a bound on the order of the hidden layer, and assuming an output $\O = \set{H}^{1}$ of degree $D=1$.
In comparison, \cref{cor:set.universality} uses a linear bound and applies to a much more general setting of arbitrary orders for the input and output product sets. In fact, the universality result is true for arbitrary input-output $\gr{S}_\I$-sets. 

\paragraph{Linear \G-Map as a Product Space}\label{sec:gprod}
For finite groups, the linear \G-map $\Phi: \Real^\I \to \Real^\O$ 
is indexed by $\O \times \I$, and therefore it is a product space. 
In fact the parameter-sharing of \cref{eq:param_sharing} ties all the parameters $\Phi(\o, \i)$ that are in  
the same orbit. 
Therefore, the decomposition \cref{eq:product_space} also identifies
 parameter-sharing pattern of $\Phi$.\footnote{When $\I$ and $\O$ are homogeneous spaces,  another characterization the orbits of the product space $[\G_\i \backslash \G] \times [\G_\o \backslash \G]$ is by showing their one-to-one correspondence with double-cosets $\G_\i \backslash \G / \G_\o = \{\G_\i \g \G_\o \mid \g \in \G\}$.}
 
 \begin{example}[\textsc{Equivariant Maps between Set Products}]
 Equation \cref{eq:product.set.decomp} gives a closed form for the decomposition of $\I^D$ into orbits.
 Assuming a similar decomposition for $\O^{D'}$, the equivariant map $\Phi: \Real^{\I^D} \to \Real^{\O^{D'}}$ is decomposed in to $\mathrm{Bell}(D + D')$ linear maps corresponding to the orbits of $\O^{D'} \times \I^D$. 
 \end{example}

\subsubsection{Burnside's Table of Marks}\label{sec:mark}
Burnside's table of marks simplifies working with the multiplication operation of the Burnside ring, and enables the analysis of \G-action on product spaces~\cite{burnside1911theory,pfeiffer1997subgroups}. 
The \emph{mark} of $\H \leq \G$ on a finite \G-set $\I$, is defined as
the number of points in $\I$ fixed by all $\h \in \H$:
\begin{align}
\fun{m}_{\I}(\H) \defeq |\{\i \in \I \mid \h \cdot \i = \i\; \forall \h \in \H\}|.
\end{align}
The interesting quality of the number of fixed points is that the total number of fixed points adds up when we add two spaces $\I_1 \cup \I_2$. Also, when considering product spaces $\I_1 \times \I_2$, any combination of points fixed in both spaces will be fixed by $\H$. This means  
\begin{align}
\fun{m}_{\I_1 \cup \I_2}(\G_i) &= \fun{m}_{\I_1}(\G_i) +  \fun{m}_{\I_2}(\G_i)\label{eq:markcup} \\
\fun{m}_{\I_1 \times \I_2}(\G_i) &= \fun{m}_{\I_1}(\G_i)  \, \fun{m}_{\I_2}(\G_i).\label{eq:marktimes}
\end{align}

Now define the \emph{vector of marks}
$\vect{m}_{\I}: \Omega(\G) \to \mathds{Z}^{n}$ as 
 \begin{align*}
 \vect{m}_{\I} \defeq [\fun{m}_{\I}(\G_1), \ldots, \fun{m}_{\I}(\G_I)]
 \end{align*}
 where $I$ is the the number of conjugacy classes of subgroups of $\G$, and we have assume a 
 fixed order on $[\G_i] \leq \G$. 
 Due to \cref{eq:marktimes,eq:markcup}, given \G-sets $\I_1,\ldots,\I_D$, we can perform elementwise addition and multiplication on the vector of integers $\vect{m}_{\I_1}, ..., \vect{m}_{\I_D}$, to obtain the mark of union and product \G-sets respectively.  Moreover, the special quality of marks, makes this vector
an \emph{injective} homeomorphism:
we can work backward from the resulting vector of marks and decompose the union/product space into homogeneous spaces.
To facilitate calculation of this vector, for any \G-set $\I$, one may use the table of marks.

\begin{table}
\caption{\small{Table of marks  $\mat{M}_\G$.}}\label{table:marks}
\scalebox{.7}{
\vspace{1em}
\begin{tabular}{r|lllllll|}
 \multicolumn{1}{c}{ } & \multicolumn{1}{c}{$\{\gr{e}\}$} & \multicolumn{1}{c}{\ldots} & \multicolumn{1}{c}{$\gr{G}_i$} & \multicolumn{1}{c}{\ldots} & \multicolumn{1}{c}{$\gr{G}_j$} & \multicolumn{1}{c}{\ldots} &  \multicolumn{1}{c}{$\G$} \\
\cline{2-8}
$\{\gr{e}\} \backslash \gr{G}$ & $|\G|$     &      &      &      &       &      &      \\
\vdots  &   \vdots  & $\ddots$    &      &      &      &      &        \\
$\G_i \backslash \G$  & $|\G : \G_i|$   &  \ldots &  $|\G : \mathrm{N}_\G(\G_i)|$    &      &       &      &      \\
\vdots  &   \vdots   &     &  \vdots      &  $\ddots$  &     &      &      \\
$\G_j \backslash \G$  &  $|\G : \G_j|$  &   \ldots   &  ${m}_{\G_j \backslash \G}(\G_i)$     &  \ldots    &    $|\G : \mathrm{N}_\G(\G_j)|$   &      &      \\
\vdots  &   \vdots   &     &   \vdots   &      &    \vdots   &   $\ddots$  &      \\
$\G \backslash \G$   &  1    & \ldots & 1    &  \ldots    &   1    &  \ldots    &    1  \\
\cline{2-8}
\end{tabular}
}
\end{table}

The \emph{table of marks} for a group $\G$, is the square matrix of marks of all subgroups on all right-coset spaces\footnote{
 $\fun{m}_{\G_i \backslash \G}(\G_j) = \fun{m}_{\G_i \backslash \G}(\g \G_j \g^{-1})$, and $\fun{m}_{\G_i \backslash \G}(\G_j) = \fun{m}_{\g \G_i \g^{-1} \backslash \G}(\G_j) \quad \forall \g \in \G$. Therefore, the table of marks' characterization is up to conjugacy.} -- that is the element $i,j$ of this matrix is:
\begin{align}
\mat{M}_\G({i,j}) \defeq \fun{m}_{\G_i \backslash \G}(\G_j) \quad \text{or} \quad
\mat{M}_\G \defeq \begin{bmatrix} 
\vect{m}_{\{e\}\backslash \G} \\
\vdots\\
\vect{m}_{\G\backslash \G}
\end{bmatrix}.
\end{align}
The matrix $\mat{M}_\G$, has valuable information about the subgroup structure of $\G$. For example, 
$\G_j$'s action on $\G_i \backslash \G$ will have a fixed point, iff $[\G_j] \leq [\G_i]$.
Therefore, the sparsity pattern in the table of marks, reflects the subgroup lattice structure of $\G$, up to conjugacy.\footnote{The sub-group lattice of $\G$ is a partially ordered set in which the order $\G_{i} < \G_{j}$ is a subgroup relation, and the greatest and least elements are $\G$ and $\{e\}$ respectively. Any \G-set is isomorphic to a right-coset space produced by a member of this lattice. However, we only care about this lattice up to a conjugacy relation.
This is because as we saw, 
the right cosets $\H \backslash \G$ and
$(\g^{-1}\H \g) \backslash \G \quad \forall \g \in \G$ are isomorphic.
}

A useful property of $\mat{M}_\G$ is that we can use it to find the marks $\vect{m}_{\I}$ on any \G-set $\I = \sum_i p_i [\G_i \backslash \G]$ in $\Omega(\G)$ using the expression 
$
\vect{m}_{\I} = [p_1,\ldots,p_I]^{\top } \mat{M}_\G.
$
Moreover, the structural constants of \cref{eq:product_space} can be recovered from the table of Marks 
\begin{align}\label{eq:table_decomposition}
\delta^\ell_{i j} = \sum_l \mat{M}_\G ({i, l}) \mat{M}_\G({j, l})  (\mat{M}_\G^{-1})({l, \ell}).
\end{align}
\begin{figure}
    \centering
    \includegraphics[width=.6\linewidth]{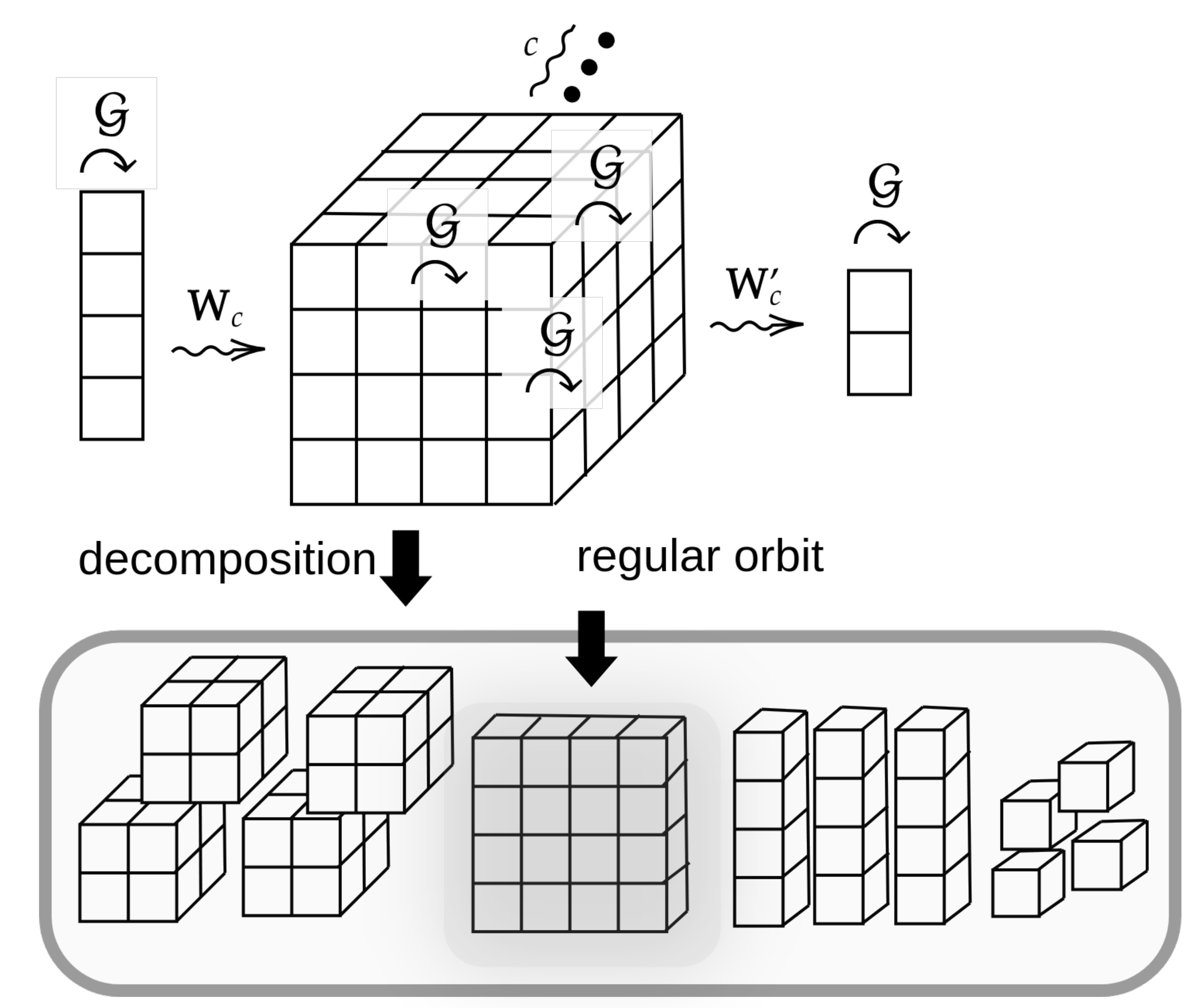}
    \caption{\footnotesize{A high-order hidden layer decomposes into orbits, which are characterized by the table of marks. By increasing the order one could guarantee the existence of a regular orbit in the decomposition. By \cref{th:universal.equivariant} this leads to universal equivariance.}}
    \label{fig:decomposition}
\end{figure}

\section{Universality of \G-Maps on  Product Spaces}\label{sec:universality_product}
Using the tools discussed in the previous section, 
in this section we prove some properties of 
product spaces that are consequential in design of equivariant maps. 
Previously we saw that product spaces decompose into orbits, identified by $\delta_{ij}^{\ell} > 0$ in \cref{eq:product_space}.
The following theorem states that such product spaces always have orbits that are at least as large as the largest of the input orbits,
and at least one of these product orbits is strictly larger than both inputs. 
For simplicity, this theorem is stated in terms of the stabilizers, rather than the orbits, where by the \emph{orbit-stabilizer} theorem, 
larger stabilizers correspond to smaller orbits.
Also, while the following theorem is stated for the product of homogeneous \G-sets,
it trivially extends to product of \G-sets with multiple orbits.

\begin{mdframed}[style=MyFrame2]
\begin{theorem}\label{th:larger}
Let $[\G_i \backslash \G]$ and $[\G_j \backslash \G]$ be transitive \G-sets, with $\{e\} < \G_i , \G_j < \G$. Their product \G-set decomposes into orbits
$[\G_i \backslash \G] \times [\G_j \backslash \G] = \bigcup_\ell \delta_{i j}^{\ell} [\G_\ell \backslash \G]$, such that:
\item (\textbf{i})  $[\G_\ell] \leq [\G_i], [\G_j]$ for all the resulting orbits.
\item (\textbf{ii}) if $\G_j \not \subseteq \mathrm{Core}_\G(\G_i)$ and 
$\G_i \not \subseteq \mathrm{Core}_\G(\G_j)$, then
$[\G_\ell] {<} [\G_i], [\G_j]$ for at least one of the resulting orbit. 
\end{theorem}
\end{mdframed}

\begin{proof}
The proof is by analysis of the table of Marks  
$\mat{M}_{\G}$.
The vector of mark for the product space is the element-wise product of vector of marks of the input:
$\vect{m}_{[\G_i \backslash \G]\times [\G_i \backslash \G]} = \vect{m}_{[\G_i \backslash \G]} \odot \vect{m}_{[\G_j \backslash \G]}.$ 
The same vector, can be written as a linear combination of rows of $\mat{M}_{\G}$, with non-negative integer coefficients:
$
\vect{m}_{\G_i \backslash \G} \odot \vect{m}_{\G_j \backslash \G} = \sum_{\ell} \delta^{\ell}_{ij} \vect{m}_{[\G_\ell \backslash \G]}.
$
For convenience we assume a \emph{topological ordering} of the conjugacy class of subgroups $\{e\} = \G_1, \ldots, \G_i, \ldots, \G_I = \G$ consistent with their partial order -- that is $[\G_i] \not > [\G_{j}] \forall j > i$.
This means that $\mat{M}_\G$ is lower-triangular, with nonzero diagonals; see \cref{table:marks}. Three important properties of this table are~\cite{pfeiffer1997subgroups}:
    (1)  the sparsity pattern in $\mat{M}_\G$ reflects the subgroup relation:
    $\vect{m}_{[\G_i \backslash \G]}(\ell) > 0$ iff $\G_\ell \leq \G_i$. 
    (2) the first column is the index of $\G_i$ in $\G$: 
    $\vect{m}_{[\G_i \backslash \G]}(1) = |\G : \G_i| \quad \forall i$.
    (3) the diagonal element is the index of the normalizer: $\vect{m}_{[\G_i \backslash \G]}(i) = |\G : N_\G(\G_i)|\; \forall i$, where the \emph{normalizer} of $\H$ in $\G$ is defined as the largest intermediate subgroup of $\G$ in which $\H$ is normal:
$N_{\G}(\H) = 
\{ 
\g \in \G \mid \g \H \g^{-1} = \H 
\}.$

(\textbf{i}) From (1) it follows that the non-zeros of the product $(\vect{m}_{[\G_i \backslash \G]} \odot \vect{m}_{[\G_j \backslash \G]})(\ell) > 0$ correspond to $\G_\ell \leq [\G_i]$ and  $\G_\ell \leq [\G_j]$. Since the only rows of $\mat{M}_\G$ with such non-zero elements are
$\vect{m}_{[\G_\ell \backslash \G]}$ for $\G_\ell \leq [\G_i] \cap \G_j$, all the resulting orbits have such stabilizers. This finishes the proof of the first claim.

(\textbf{ii})  
If $[\G_i] \not \leq [\G_j]$ and $[\G_j] \not \leq [\G_i]$, then $[\G_\ell]$ which is a subgroup of both groups is strictly smaller than both, which means one of the resulting orbits must be larger than both input orbits.
Next, w.l.o.g., assume $[\G_i] \leq [\G_j]$. Consider proof by contradiction: suppose the product does not have a strictly larger orbit. It follows that
$
\vect{m}_{[\G_j \backslash \G]} \odot \vect{m}_{[\G_i \backslash \G]} = \delta^{i}_{i,i} \vect{m}_{[\G_i \backslash \G]}
$ 
for some $\delta_{ii}^{i} > 0$.
Consider the first and $i^{th}$ element of 
the elementwise product above:
\begin{align*}
|\G : \G_j| \times |\G : \G_i| &= \delta^{i}_{i i} |\G : \G_i| \\
\vect{m}_{[\G_j \backslash \G]}(i) \times |\G : N_\G(\G_i)| &= \delta^{i}_{i i} |\G : N_\G(\G_i)|
\end{align*}
Substituting $\delta_{i i}^i = |\G : \G_j|$
from the first equation into the second 
equation and simplifying we get
$
\vect{m}_{[\G_j \backslash \G]}(i) = |\G : \G_j|.
$
This means the action of $\G_i$ on $[\G_j \backslash \G]$
fixes all points, and therefore $\G_i \subseteq \mathrm{Core}_\G(\G_j)$ as defined in \cref{eq:core}.
This contradicts the assumption of (ii).
\end{proof}

\begin{table}
\caption{\small{Table of marks for the alternating group $\gr{A}_5$.}}\label{table:a5}
\scalebox{.7}{
\vspace{1em}
\begin{tabular}{r|lllllllll|}
\multicolumn{1}{c}{}& \multicolumn{1}{c}{$\{\gr{e}\}$} & \multicolumn{1}{c}{$\gr{C}_2$} & \multicolumn{1}{c}{$\gr{C}_3$} & \multicolumn{1}{c}{$\gr{K}_4$} & \multicolumn{1}{c}{$\gr{C}_5$} & \multicolumn{1}{c}{$\gr{S}_3$} & \multicolumn{1}{c}{$\gr{D}_{10}$} & \multicolumn{1}{c}{$\gr{A}_4$} & \multicolumn{1}{c}{$\gr{A}_5$}\\
\cline{2-10}
$\{\gr{e}\} \backslash \gr{A}_5$ &   60  &      &      &      &      &      &       &      &      \\
$\gr{C}_2 \backslash \gr{A}_5$  & 30    & 2    &      &      &      &      &       &      &      \\
$\gr{C}_3 \backslash \gr{A}_5$  & 20    &      & 2    &      &      &      &       &      &      \\
$\gr{K}_4 \backslash \gr{A}_5$  & 15    & 3    &      & 3    &      &      &       &      &      \\
$\gr{C}_5 \backslash \gr{A}_5$  & 12    &      &      &      & 2    &      &       &      &      \\
$\gr{S}_3 \backslash \gr{A}_5$  & 10    & 2    & 1    &      &      & 1    &       &      &      \\
$\gr{D}_{10} \backslash \gr{A}_5$ & 6     & 2    &      &      & 1    &      & 1     &      &      \\
$\gr{A}_4 \backslash \gr{A}_5$  & 5     & 1    & 2    & 1    &      &      &       & 1    &      \\
$\gr{A}_5 \backslash \gr{A}_5$  & 1     & 1    & 1    & 1    & 1    & 1    & 1     & 1    & 1   \\
\cline{2-10}
\end{tabular}
}
\end{table}   

A sufficient condition for (ii) in \cref{th:larger} is for the \G-action on input \G-sets to be faithful. Note that in this case the the core is trivial; see \cref{sec:classification}. 
An implication of this theorem is that repeated self-product $[\H \backslash \G]^{D}$ is bound to produce a regular orbit.
This leads to \cref{cor:degree}, that we saw earlier. Here, we give a shorter proof using 
\cref{th:larger}; see \cref{fig:decomposition}.
\begin{proof}[Alternative Proof of \cref{cor:degree}]
Since \G acts faithfully on $\I$, $\mathrm{Core}_\G(\H) = \{\gr{e}\}$.  
From \cref{th:larger} it follows that each time we calculate a product by $\I$, a strictly smaller stabilizer is produced so that 
$\H = \H^{(t=0)} > \H^{(1)} > \ldots > \H^{(D)} = \{e\}$, where $\H^{(d)}$ is the smallest stabilizer at time-step $d$. From Lagrange theorem, the size of a proper subgroup is at most half the size of its overgroup in this sequence of stabilizers. It follows that for any $D \geq \log_2 |\H|$, $[\H \backslash \G]^{D}$ has an orbit with $\H^{t=D} = \{e\}$ as its stabilizer. 
\end{proof}

\begin{example}[\textsc{Universal Approximation for $\gr{A}_5$}]
The alternating group $\gr{A}_5$ is the group of even permutations of 5 objects.
One way to create a universal approximator for this group to have a regular layer (see \cref{th:universal.equivariant}). 
A more convenient alternative is to consider the canonical 
action of this group on a set $\I$ of size $5$, and use an order 
$D$ layer to ensure universality.
Using \cref{cor:loose} we get 
$D \geq 5 = \lceil (3\frac{1}{2} \log_2 (4))  - 4 \log_2(e) \rceil.$
The natural action of $\gr{A}_{5}$ on $\set{N} = [5]$ is isomorphic to $[\gr{A}_4 \backslash \gr{A}_5]$ -- \ie $\gr{A}_4$ is a stabilizer. Using this stabilizer in
 \cref{cor:degree}, we get the same bound
$D \geq  5  = \lceil \log_2(|\gr{A}_4|) \rceil.$

However, 
using the table of marks we can show that $D=3$ already
produces a regular orbit in this case.
The table of marks for the alternating group $\gr{A}_5$ is shown in \cref{table:a5}. Our objective is to find the decomposition of $[\gr{A}_4 \backslash \gr{A}_5]^3$. 
We do this in steps, first showing 
\begin{align}\label{eq:a5_1}
    [\gr{A}_4 \backslash \gr{A}_5]^2 = [\gr{A}_4 \backslash \gr{A}_5] \cup [\gr{C}_3 \backslash \gr{A}_5]
\end{align} 
To see this, note that the element-wise product of the 
vector of marks $\vect{m}_{[\gr{A}_4 \backslash \gr{A}_5]}$ (which is next to last row in \cref{table:a5}) with itself
is equal to 
$\vect{m}_{[\gr{A}_4 \backslash \gr{A}_5]} + \vect{m}_{[\gr{C}_3 \backslash \gr{A}_5]}$. Since the vector of marks is an injective homomorphism, this implies \cref{eq:a5_1}. Applying the same idea one more time, gives
\begin{align*}
    [\gr{A}_4 \backslash \gr{A}_5]^3 = ([\gr{A}_4 \backslash \gr{A}_5] \cup [\gr{C}_3 \backslash \gr{A}_5]) \times [\gr{A}_4 \backslash \gr{A}_5] \\
= 2[\gr{A}_4 \backslash \gr{A}_5] \cup [\gr{C}_3 \backslash \gr{A}_5] \cup [\{e\} \backslash \gr{A}_5].
\end{align*}

This shows that $[\gr{A}_4 \backslash \gr{A}_5]^3$ contains a regular orbit $[\{e\} \backslash \gr{A}_5]$. Therefore, using an order $D=3$
hidden layer $\I^3$ on which $\gr{A}_5$ acts using even permutations, 
also produces a universal equivariant (invariant) approximator.
\end{example}

\section*{Acknowledgements}
We thank anonymous reviewers for their constructive feedback. In particular the first proof for \cref{cor:degree}, as well as clarifications on the proof of the main theorems was proposed by reviewers.
This research is in part funded by the Canada CIFAR AI Chair Program.
\nocite{ravanbakhsh_sets,sannai2019universal}
\bibliography{references.bib}
\bibliographystyle{icml2020}

\end{document}

%% file: math_commands.tex
\newtheorem{corollary}{Corollary}
\newtheorem{example}{Example}

\usepackage[framemethod=TikZ]{mdframed}
\mdfdefinestyle{MyFrame}{%
    linecolor=black,
    outerlinewidth=.3pt,
    roundcorner=5pt,
    innertopmargin=1pt, 
    innerbottommargin=1pt, 
    innerrightmargin=1pt,
    innerleftmargin=1pt,
    backgroundcolor=black!0!white}

\mdfdefinestyle{MyFrame2}{%
    linecolor=white,
    outerlinewidth=1pt,
    roundcorner=2pt,
    innertopmargin=\baselineskip,
    innerbottommargin=\baselineskip,
    innerrightmargin=10pt,
    innerleftmargin=10pt,
    backgroundcolor=black!3!white}

\newcommand{\ie}[0]{\emph{i.e.},~}

\newcommand{\aka}[0]{a.k.a.~}

\newcommand{\defeq}{\ensuremath{\doteq}}

\newcommand{\mat}[1]{\ensuremath{{\mathbf{\MakeUppercase{{#1}}}}}}
\newcommand{\fun}[1]{\ensuremath{{{#1}}}}
\newcommand{\vect}[1]{\ensuremath{{\mathbf{\MakeLowercase{{#1}}}}}}
\newcommand{\set}[1]{\ensuremath{\mathbb{#1}}}
\newcommand{\setb}[1]{\ensuremath{\mathbb{#1}}}
\newcommand{\gr}[1]{\ensuremath{\mathcal{#1}}}

\newcommand{\Real}{\mathds{R}}

\renewcommand{\S}{\gr{S}}
\newcommand{\G}{\gr{G}\xspace}
\newcommand{\g}{\gr{g}}
\renewcommand{\H}{\gr{H}}
\newcommand{\h}{\gr{h}}
\newcommand{\K}{\gr{K}}

\newcommand{\I}{\setb{N}}
\renewcommand{\i}{n}
\renewcommand{\O}{\setb{M}}
\renewcommand{\o}{m}
\renewcommand{\Phi}{\mat{W}}
\newcommand{\Phis}{\Phi}
\renewcommand{\Psi}{\mat{L}}
\renewcommand{\varphi}{{w}}
\newcommand{\x}{\vect{x}}